\documentclass[12pt]{article}

\usepackage[T1]{fontenc}
\usepackage{mathptmx}

\usepackage{newtxtext,newtxmath}

\usepackage[utf8]{inputenc}
\usepackage{graphicx}
\usepackage{caption}
\usepackage{amsmath}
\usepackage[utf8]{inputenc}
\usepackage{hyperref}
\usepackage{graphicx}
\usepackage{multirow}
\usepackage{amsmath}
\usepackage[left=15mm,right=15mm,top=20mm,bottom=20mm]{geometry}
\usepackage[gen]{eurosym}
\usepackage{pdfpages}
\usepackage{amsmath}
\usepackage{graphicx}
\usepackage{float}
\usepackage{subcaption}
\usepackage{gensymb}
\usepackage{caption}
\usepackage{subcaption}
\usepackage{pdfpages}
\usepackage{nth}
\usepackage{subcaption}

\usepackage{sectsty}
\sectionfont{\large\bfseries}

\usepackage{ragged2e}
\usepackage[utf8]{inputenc}
\usepackage{hyperref}
\usepackage{graphicx}
\usepackage{multirow}
\usepackage{amsmath}
\usepackage[left=15mm,right=15mm,top=20mm,bottom=20mm]{geometry}
\usepackage[gen]{eurosym}
\usepackage{pdfpages}
\usepackage{amsmath}
\usepackage{graphicx}
\usepackage{float}
\usepackage{subcaption}
\usepackage{gensymb}
\usepackage{caption}
\usepackage{subcaption}
\usepackage{pdfpages}
\usepackage{nth}
\usepackage{listings}
\usepackage{color} 
\usepackage{abstract}

\definecolor{mygreen}{RGB}{28,172,0} 
\definecolor{mylilas}{RGB}{170,55,241}
\begin{document}

\begin{center}

\parindent=0in \newcommand{\HRule}{\rule{\linewidth}{0.5mm}}


\vspace*{5mm}

\HRule \\[0.4cm] { \Large{\textbf{A Bioinspired Stiffness Tunable Sucker for Passive Adaptation and Firm Attachment to Angular Substrates} }}\\[0.3cm] \HRule \\[1.5cm]

\begin{minipage}{\textwidth}\large

\begin{center}
\begin{tabular}{c c}
\begin{minipage}[t]{3in}
\normalsize
\centering
\textbf{Arman Goshtasbi}\\
Department of Biomechanical Engineering\\
University of Twente\\
Enschede, The Netherlands\\
a.goshtasbi@student.utwente.nl
\end{minipage}
&
\begin{minipage}[t]{3in}
\centering
\normalsize
\textbf{Ali Sadeghi}\\
Department of Biomechanical Engineering\\
University of Twente\\
Enschede, The Netherlands\\
a.sadeghi@utwente.nl

\end{minipage}
\end{tabular}
\end{center}

\end{minipage}

\end{center}

\vspace{2cm}

\begin{abstract}
\normalsize
\noindent The ability to adapt and conform to angular and uneven surfaces improves the suction cup's performance in grasping and manipulation. However, in most cases, the adaptation costs lack of required stiffness for manipulation after surface attachment; thus, the ideal scenario is to have compliance during adaptation and stiffness after attachment to the surface. Nevertheless, most stiffness modulation techniques in suction cups require additional actuation. This article presents a new stiffness tunable suction cup that adapts to steep angular surfaces. Using granular jamming as a vacuum driven stiffness modulation provides a sensorless for activating the mechanism. Thus, the design is composed of a conventional active suction pad connected to a granular stalk, emulating a hinge behavior that is compliant during adaptation and has high stiffness after attachment is ensured. During the experiment, the suction cup can adapt to angles up to 85$^\circ$ with a force lower than 0.5 N. 
We also investigated the effect of granular stalk's length on the adaptation and how this design performs compared to passive adaptation without stiffness modulation.
\end{abstract}

\section{Introduction}

Adaptation plays a pivotal role in grasping, manipulating, and interacting with unknown environments. Due to their low mechanical flexibility, conventional robots require complex control and sensory and actuation systems to achieve such adaptation \cite{soft_manipulator_and_gripper}. In most cases, these robots lack adaptability and are only suitable for a single task \cite{Desing_fab}. On the other hand, inspired by animals' inherent ability to adapt to unknown surroundings, studies suggest that soft grippers, thanks to their compliance, can exceed rigid robots' limitations and perform more adaptively in undefined conditions \cite{adapt}. 

Many animals utilize astrictive prehension, also known as adhesive grippers, to attach to surfaces or grab objects. Different techniques have been suggested to achieve adhesion inspired by such animals. For instance, different grippers have been developed using van der Waals force inspired by microfibers on gecko's toes \cite{gecko_cut}, \cite{gecko_song}. However, controlling the force in such grippers is difficult \cite{gecko_3}. Another technique, inspired by suction cups in octopus tentacles \cite{octo}, and suction disc on northern clingfish ventral side \cite{Cling_2019} \cite{cling2}, is surface attachment using negative fluid pressure in a suction cup. Due to their fast, controllable, and efficient way of attaching to surfaces \cite{Follador_2014}, vacuum suction grippers have shown great potential in various grasping applications, such as surface grasping in wall-climbing robots \cite{wall_1}, grasping in surgical application \cite{surgic}, and haptic exploration \cite{haptic}. The suction cups are either passive, in which the change in the shape provides negative pressure, or active, in which an external vacuum pump generates the pressure. The active vacuum gripper produces more grasping force and is easier to detach by connecting it to ambient pressure. At the same time, the latter provides a tetherless solution and a more energy-efficient solution for grasping \cite{Follador_2014}.  

Despite the numerous advantages of suction cups, conventional suction cups have some limitations in adaptation to various environments. For instance, rough and porous surfaces prevent the vacuum suction cups from providing negative pressure. Several studies have proposed designs to overcome these shortcomings by looking at how animals overcome such problems. For example, Inspired by sea urchins, combination of soft suction pad and chemical adhesive material improve the grasping on very rough surfaces (\cite{Seaurchin}). \cite{Baik2017AWA} microfabricated a thin elastomer with multiple micro-suction cups in an octopus-like suction cup. Furthermore, \cite{Takahashi}, Developed a micro-bumps surface for rough surface grasping. Aside from the bio-inspired designs, \cite{koivikko} designed a filmed base suction cup that can attach to rough and porous surfaces.

Another challenge for vacuum suction cups is grasping objects with unknown-shaped objects and angular surfaces. The suction cups require flexibility to adapt to the object's surface to overcome this issue. \cite{Song} suggest attaching a thin elastomer membrane to the suction cup, enabling it to attach to round objects. \cite{zhakypov} proposed an origami-based design to adjust to the object's shape by actuating shape memory alloys. Finally, inspired by Octopus, \cite{mazzolai_stalk} attached the suction cups to a flexible stalk to emulate spherical joint behavior.

Although each design solves a critical problem, most of them pay the most attention to grabbing an object rather than attaching it to an angular surface, which can benefit applications such as manipulating unknown objects. Furthermore, the flexibility in these designs that makes the adaptation possible causes a lack of stiffness, which can be problematic for manipulation applications. Several researchers have investigated methods to tackle this paucity using stiffness modulation techniques, such as employing layer jamming to increase and control the vertical stiffness, enabling lifting heavier objects \cite{layer_jamming}. In addition, \cite{Kim} designed a cloth-rubber beam to provide enough stiffness for surgical applications. Also, in the origami design of \cite{zhakypov}, the shape memory alloys endow the design with the possibility to lift 50 times more of its weight. However, most studies do not discuss bending stiffness, which plays a significant role in wall-climbing applications. 

In nature, octopus suction cups provide the best solution for adaptation and bending stiffness. Octopus, similar to other Cephalopod mollusks, such as squids and cuttlefish, employ muscular hydrostats which make them capable of active stiffening and force generation \cite{muscular_hydrostats}\cite{external_muscle}. In the suction cup, the octopus uses muscular hydrostats to not only adapt the surface and generate the negative pressure required for grasping but also control the bending stiffness and manipulate the grasped object using the extrinsic muscle of the suction cup. The extrinsic muscle is the muscle that connects the sphincter to the arm of the octopus. 

In this study, we propose a new granular jamming base suction cup inspired by the rotary behavior of octopus suction cups \cite{mazzolai_stalk}\cite{external_muscle} and jamming designs \cite{hinge} \cite{Ranzanijamming}. This design adapts steep angular surfaces with the material's compliance and provides high bending stiffness after the attachment with the vacuum of granular particles. The stiffness modulation in this design does not require an additional actuator, which helps weight and energy efficiency. In addition, the design is sensorless and simplifies the control in many applications. This study investigated how changing parameters such as pressure, angle, and material compliance affect the design performance. 

\section{Material and Methods}

\subsection{Design and Fabrication}

\begin{figure}[h]
    \centering
    \includegraphics[width=1\textwidth]{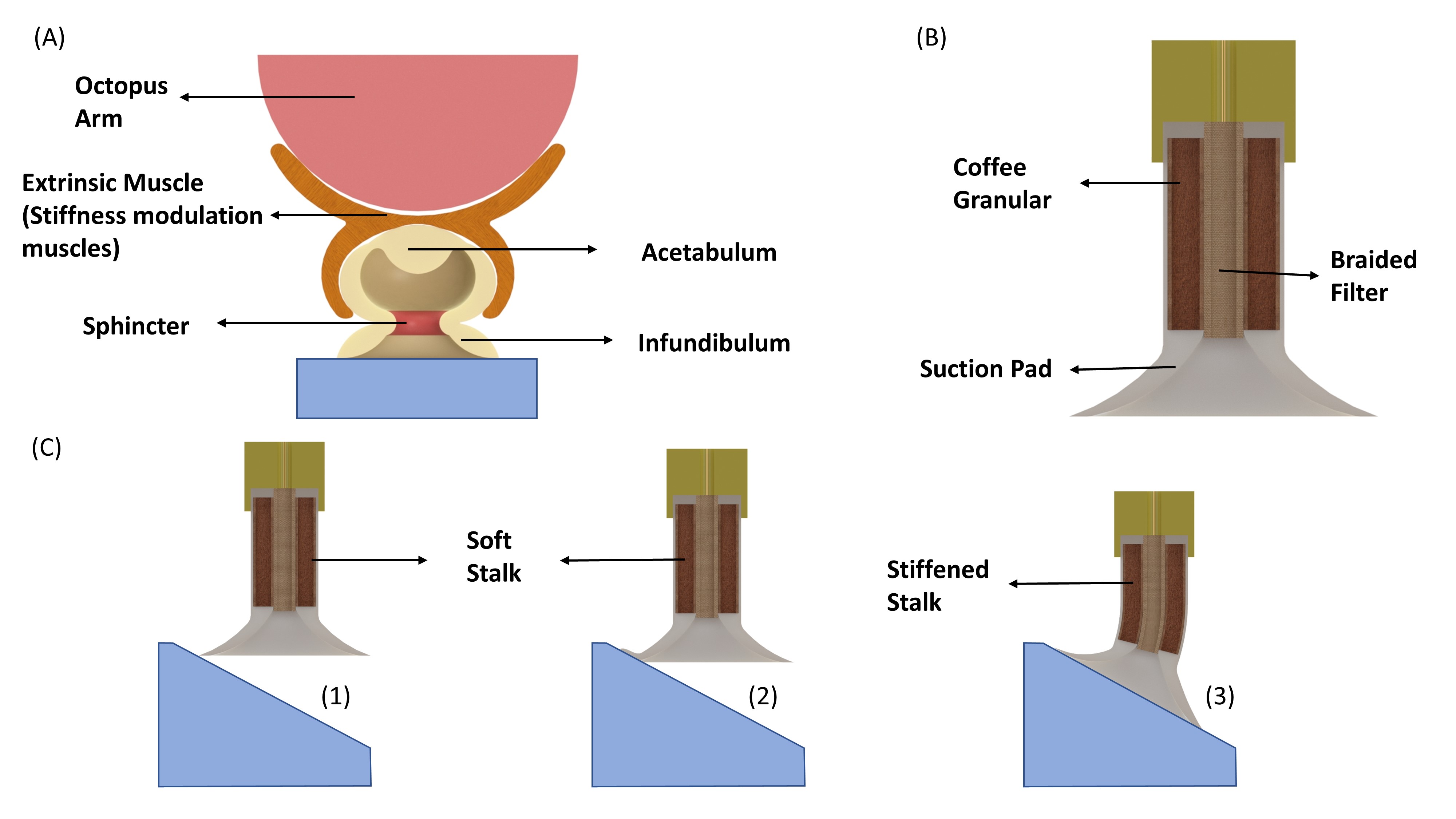}
    \caption{(A) Schematic drawing of octopus suction cup and the connection to arm via extrinsic muscle. The extrinsic muscle enable the octopus suction cup to have rotary manipulation and also provide stiffness modulation. (B) Drawing of the proposed suction cup. (C) The adaptation sequence of the suction cup to a angular surface. 
    }
    \label{fig:design}
\end{figure}

In the design shown in Fig \ref{fig:design}, a soft stalk is attached to a conventional suction pad to imitate wrist-like behavior for passive adaptation to angular surfaces. Although in this design, the flexibility of compliant materials renders the possibility of adaptation to steep angles, the low stiffness of such materials reduces the bending tolerance which can be disadvantageous in some applications such as wall-climbing and grasping object from lateral side. Therefore, stiffness modulation techniques endow this design with high flexibility of compliant materials for adaptation and high shear resistance after surface attachment. We choose granular jamming as a stiffness modulation method, since it can be activated by negative pressure. This creates the possibility to use the same source of negative pressure for activation of both suction cup and stiffness tunable stalk. This stalk design perform similar to octopus extrinsic muscle which enables angular manipulation and adaptation for the suction cup. 

\begin{figure}[h]
    \centering
    \includegraphics[width=1\textwidth]{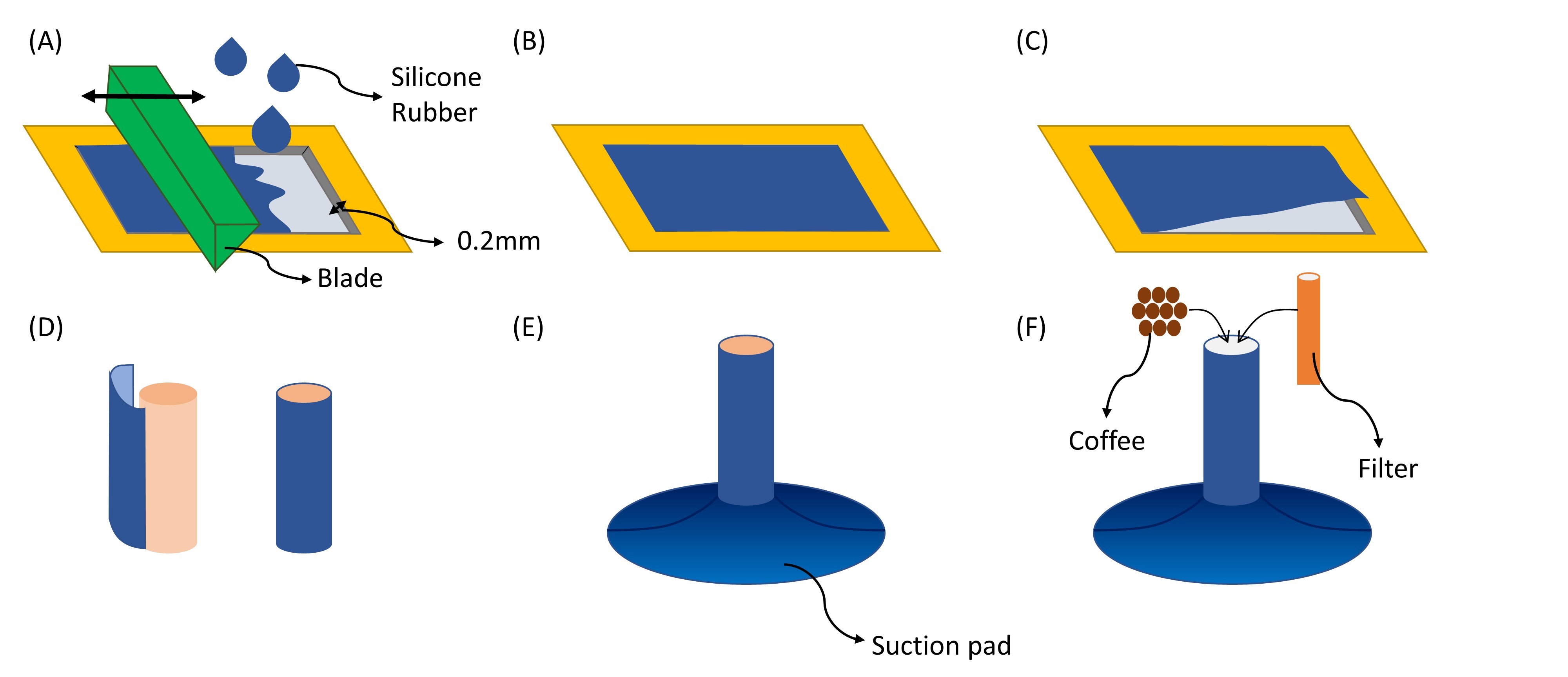}
    \caption{(A) Casting silicone rubber on a plate and use blade to make uniform thin film. (B) The silicone rubber cast on plate and left to b cured. (C) Peeling the film from plate (D) Wrapping the film around a cylinder and sealing two sides of the film. (E) Attaching the suction pad to the stalk. (F) Adding coffee and place filter in the suction cup.}
    \label{fig:fabrication}
\end{figure}

For fabrication of the suction pad, we degassed silicone rubber (1atm, degassing time: 10min) and poured it into a 3D-printed mold. For the experiments, series of suction cups were fabricated with different silicone rubbers (Ecoflex 00-10 and Dragonskin 10, Smooth ON) to study how the suction pad softness affects the adaptation. In the fabrication of the soft stalk, as presented in Fig\ref{fig:fabrication}, we cast a thin film of Ecoflex 00-10 with a thickness of 0.2 mm on a plate using the doctor blade casting method \cite{doctorblade} and cured it in the oven (60$^\circ$). Using such a thin film makes the design significantly compliant during adaptation and stiff after attachment. After the silicon was cured, we peeled off the film and rounded it on a cylinder with a diameter of 12mm to attach two sides of the film using the same silicone rubber. Finally, the suction cup was cast on top of the cylindrical film. Based on \cite{granular_test}, the area in the thin film tube was filled with coffee grains as the granular particle to enhance modulation. In addition to the granular stalk, we cast silicon rubber arms (Ecoflex 00-10 and Dragonskin 10) with the exact dimensions as the granular stalk (20mm) to compare the adaptation of the proposed design with adaptation using only material flexibility and without stiffness modulation. As shown in Fig\ref{fig:design}, a fine braided textile was used as cylindrical filter and is placed in the middle of the suction cup to have an air connection and prevent coffee grains from entering the pump. In addition to preventing coffee from going into the pump, the filter also improves the suction cup's axial load capacity.

\subsection{Theory}

\begin{figure}[h]
    \centering
    \includegraphics[width=0.7\textwidth]{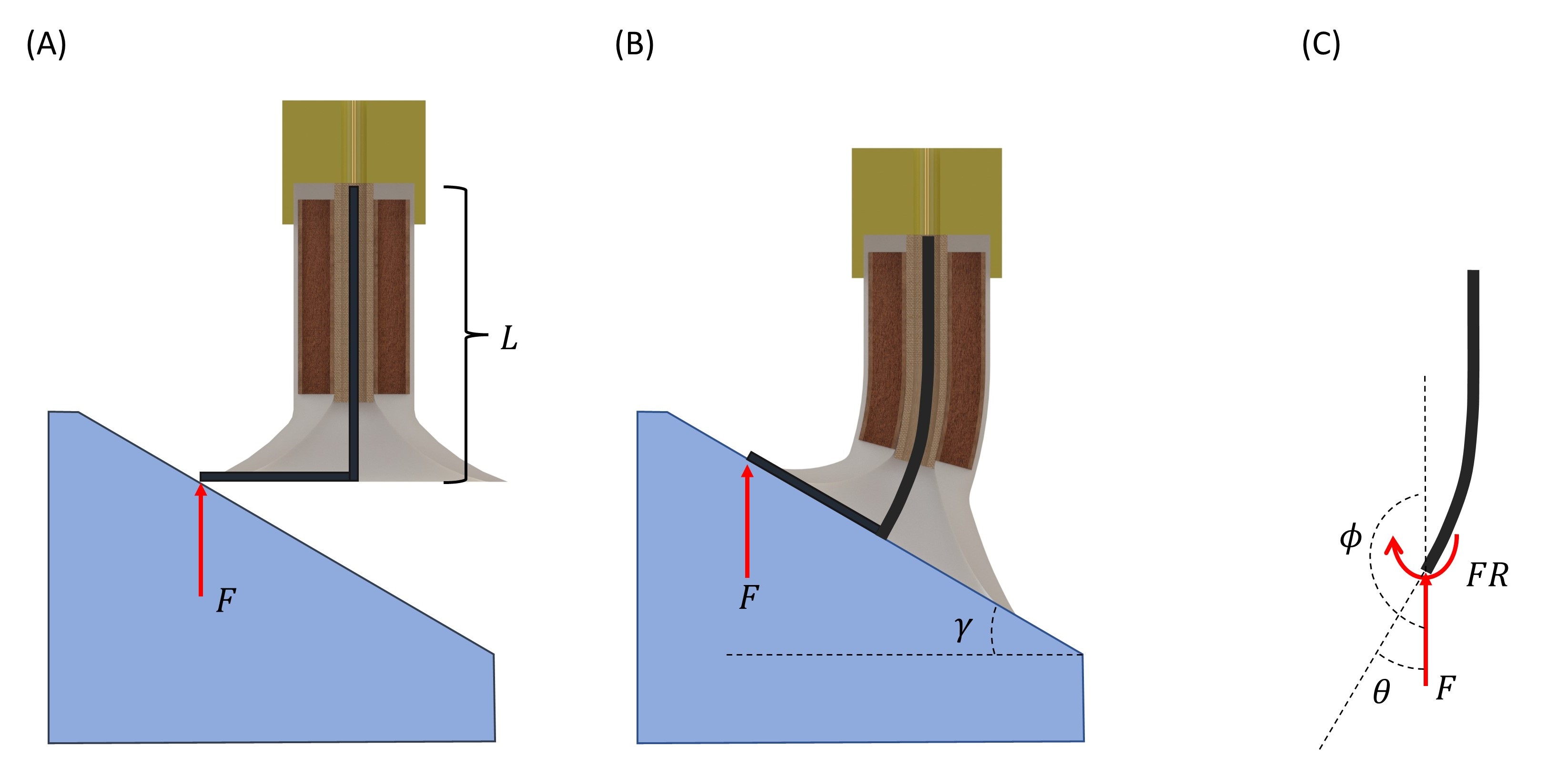}
    \caption{The L-beam used to model the behavior of the suction cup during the adaptation. (A) the initial shape of the beam and the suction cup. (B) The deformed suction cup with the deformed L-shape beam. (C) the free body diagram (FBD) of the stalk, here $F$ is the required force for adaptation, $\gamma$ is the surface angle, $\theta$ is the tip deflection, $\phi$ is the angle between the force and stalk which in this case is $180^\circ$, and also a moment relative to the force and the radius of the suction pad is also applied to the beam.}
    \label{fig:theory}
\end{figure}

As shown later, the dynamic behavior of the suction cup during the adaptation is complex, especially with granular particles moving during the deformation. Therefore, we modeled the suction pad and the stalk as an L-shape beam to simplify such complicated behavior, as presented in Fig\ref{fig:theory}. Since the stalk deforms significantly during the adaptation, the conventional static beam load-deflection equations are invalid. Hence, the dynamic beam deflection model presented in \cite{howell_2001} is used. In this model, when a straight beam is under load F with force angle $\phi$, and moment M, the beam bending differential equation is 

\begin{equation}
    \frac{d^2 \theta(s)}{ds^2} = \frac{Fsin(\theta(s)-\phi)}{EI} \quad with \quad \theta(0) = 0 \quad \theta(L) = \frac{M}{EI}
    \label{eq:1st}
\end{equation}

Where E is Young’s modulus, I is the second moment of inertia, and S is the length coordinate which is between 0 and the length of the beam (L). In our case, as shown in the free body diagram (FBD) in Fig\ref{fig:theory}-C, the moment depends on the force applied to the stalk. In addition, Eq\ref{eq:1st} is dependent on geometry and material properties. Therefore, to make the equation independent of the length of the stalk and the stalk material, Eq\ref{eq:1st} is normalized as follows:

\begin{equation}
    \frac{d^2 \theta(s)}{ds^2} = \alpha sin(\theta(\bar{s})-\phi) \quad with \quad \theta(0) = 0 \quad \frac{d\;\theta(L)}{ds} = \frac{\alpha R}{L}  \quad Define: \alpha = \frac{FL^2}{EI}
    \label{eq:2nd}
\end{equation}

Where R is the radius of the suction pad. This equation is the second order differential equation with two boundary condition. In order to solve Eq\ref{eq:2nd}, we used a numerical approach presented in \cite{numerical_integration}. To find the $\alpha$ for each angle, we numerically solve the equation until the tip angle ($\theta$) equals the surface angle ($\gamma$). By solving this equation, the $\alpha$ required for adaptation to various angles can be calculated, as shown in Table\ref{table:alpha} for angles from $0^\circ$ to $90^\circ$ with $15^\circ$ increments. Later, to convert the calculated $\alpha$ to the adaptation force, we used measured data from the stiffness test to calculate the stiffness of each scenario.

\begin{table}[h]
\centering
\caption{The required $\alpha$ for different angles from the theory presented in Eq\ref{eq:2nd}}
\vspace{2mm}
\renewcommand{\arraystretch}{1.2}
\begin{tabular}{ |c|c| } 
\hline
 Angle($^\circ$)  & Required $\alpha$ \\[0.5ex]
 \hline
  0 & 0  \\ 
  \hline
  15 & 0.445 \\
 \hline
  30 & 0.772 \\
 \hline
  45 & 1.03 \\
 \hline
  60 & 1.254 \\
 \hline
  75 & 1.467 \\
 \hline
\end{tabular}
\renewcommand{\arraystretch}{1}
\label{table:alpha}
\end{table}

\subsection{Experimental Setup}
\label{experimental_setup}
\begin{figure}[h]
    \centering
    \includegraphics[width=0.8\textwidth]{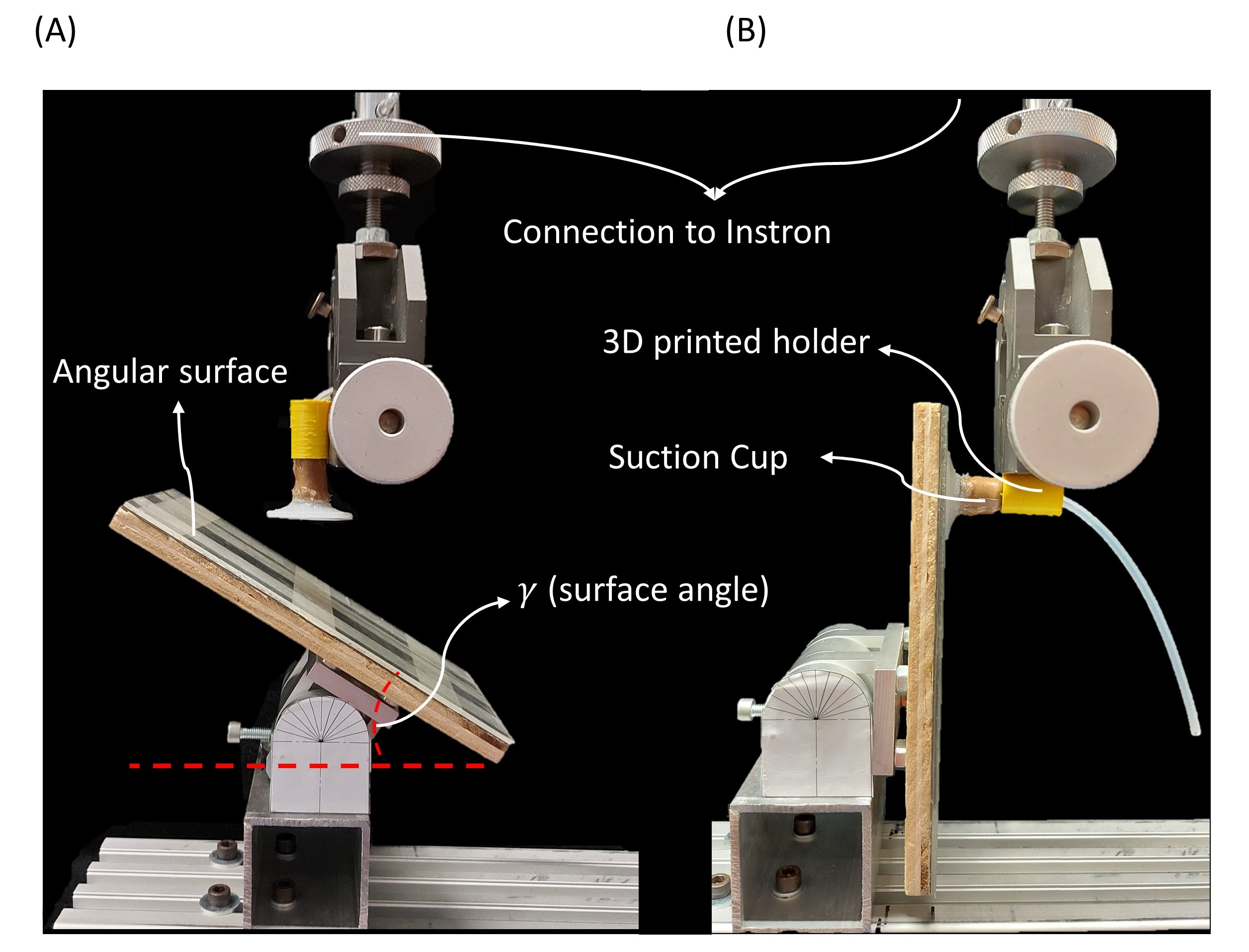}
    \caption{The setup used for adaptation and bending stiffness test. (A) the adaptation test setup:  (B) The bending stiffness test setup}
    \label{fig:setup}
\end{figure}

As mentioned, the design goal is high flexibility during adaptation and high stiffness after attachment. We performed two experiments to characterize the design behavior for both claims. A universal tensile test machine (Instron 3343, Instron, USA) with the shown setup (Fig \ref{fig:setup}) was used for all the tests. In the first experiment, the adaptation force of the suction cup to the angular surface was measured by vertically pushing the suction pad to the surface with constant speed (1mm/sec) while a vacuum pump provided the negative pressure. The applied force was measured until the suction pad was attached to the surface, monitored by measuring the pressure sensor. Then, we experimented with various surface angles (15$^{\circ}$ to 90$^{\circ}$ with 15$^{\circ}$ increments) to measure adaptation force and ultimate adaptation angle. For each scenario, the test was conducted 4 times to check the repeatability of the suction cup behavior.

Furthermore, we conducted the experiment with different lengths of the granular stalk (5-10-20mm) and different stalk materials (Ecoflex 00-10 and Dragonskin 10) but with the same diameter to exhibit how stiffness change affects the adaptation force and ultimate adaptation angle, and also how the granular joint performs compare to uniform soft material. Finally, to investigate the effect of suction pad material on adaptation, an Ecoflex 00-10 suction pad was attached to the 20mm granular joint, and the adaptation test was performed. In all adaptation tests, the surface was plexiglass, and the pressure after grasping was -60KPa relative pressure. 

We performed the bending stiffness experiment with the setup shown in Fig \ref{fig:setup} to assess the stiffness modulation impact on the bending stiffness of the stalk.  In this experiment, while the suction pad was attached to the surface with the vacuum pump, the tensile test machine applied a force to the end of the soft stalk with a constant speed of (1mm/sec). The applied force to the stalk end and the respective deflection was measured until the deflection was 5mm. To evaluate the effect of the length factor on the stalk stiffness, we carried out the test with different joint lengths (same as the previous experiment). Moreover, the bending stiffness of silicone rubber joints from earlier experiments was measured to compare to granular joint performance. In this test, the measured force demonstrates how much lateral force the suction cup tolerates in the worst situation before deforming 5mm.


\section{Result}
\subsection{Adaptation Test}

\begin{figure}[h]
\centerline
{\includegraphics[width=1\textwidth]{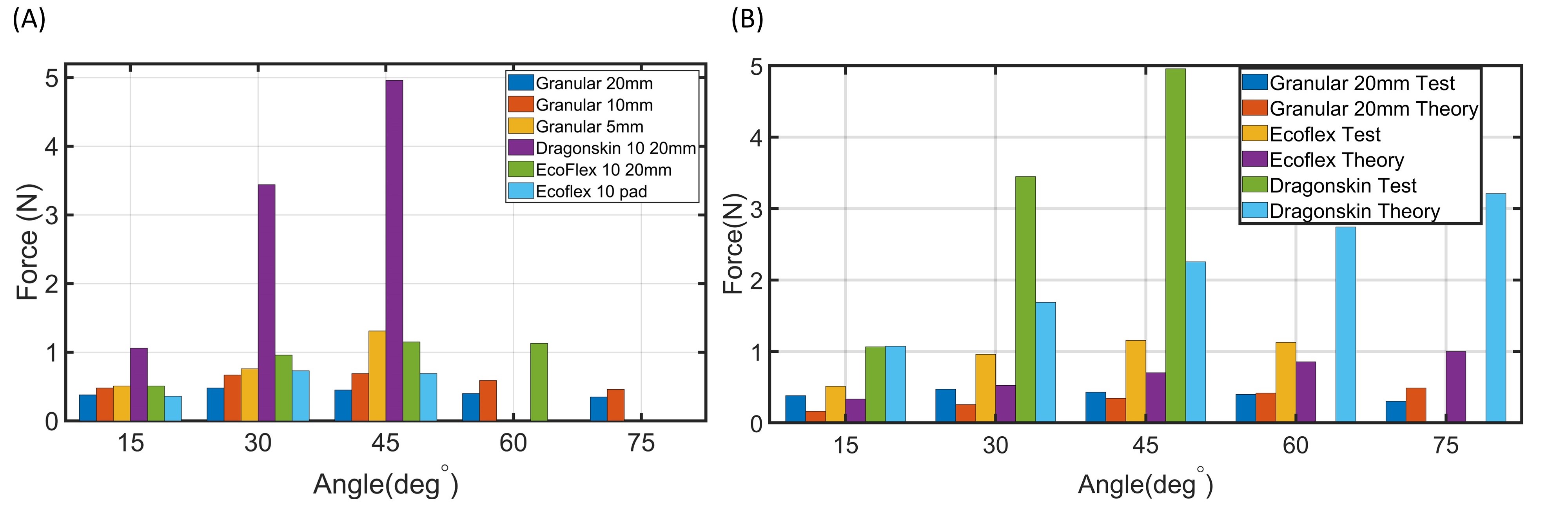}}
\caption{(A) Measured required force for adaptation to various angles. Since at 60$^{\circ}$ and 75$^\circ$ 5mm granular, Dragonskin 10 and Ecoflex 10 pad suction cup could not attach to surface, the values are not shown. (B) The measured value of the adaptation force and the required force calculated from the theory. The stiffness of different cases was calculated using the measurements shown in Fig\ref{fig:bending}. Similar to (A), the missing data are for the cases that suction cup could not attach.} 
\label{fig:adapatation test}
\end{figure}

\begin{figure}
\centerline
{\includegraphics[width=1\textwidth]{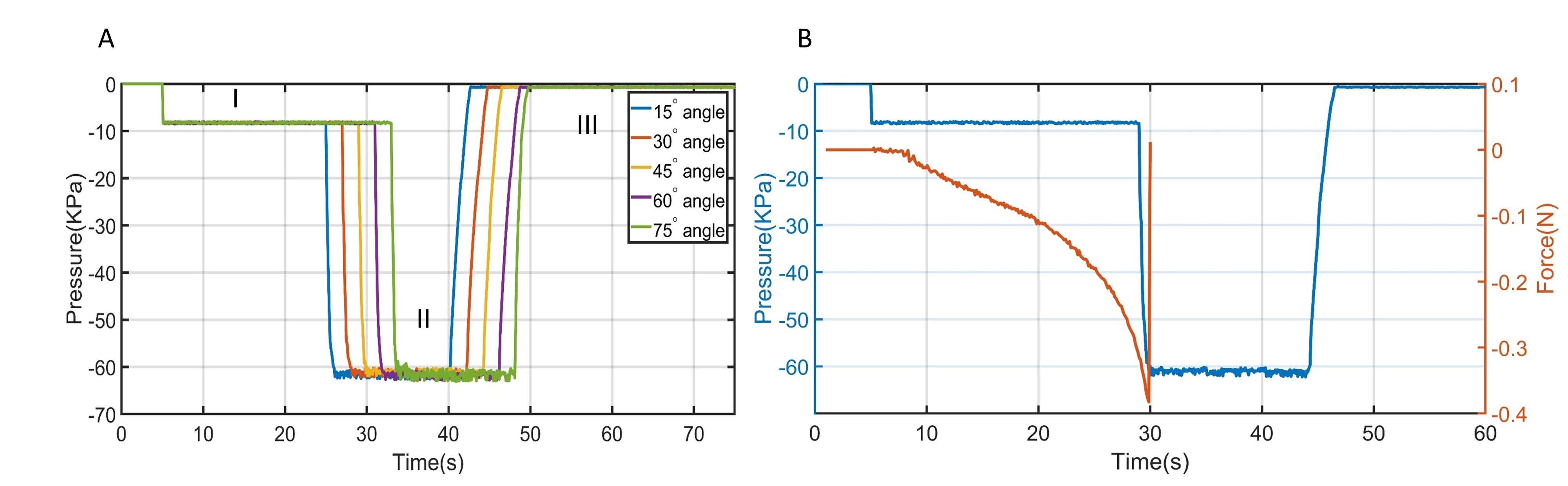}}
\caption{(A) The recorded pressure during the adaptation test for various angles (I) the initial pressure due to the self jamming of the granular particle during the adaptation (II) The pressure after the surface attachment (III) the pressure after detachment. (B) The pressure against the adaptation force at 45$^\circ$ surface angle. The force is only recorded until the suction cup attached to the surface, since after the attachment the tensile tester record the surface weight.} 
\label{fig:pressure_vs_force}
\end{figure}

In this test, we experimented with the adaptation of the proposed design to various angles and measured the required force for such adaptation. We performed this experiment with different lengths of the granular joint, joint materials, and suction pad materials. As mentioned, each prototype was tested four times to check the repeatability of the suction cup performance. In general, as presented in Fig \ref{fig:adapatation test} and Table \ref{table}, the adaptation experiment demonstrated that the proposed suction cup could adapt to very steep angles with a small force. For instance, with a 20mm granular joint, the suction cup was adapted to various angles with a force range of 0.33N at 85$^\circ$ to 0.48N at 30$^\circ$ and had ultimate adaptation angles of 85$^\circ$ (See Supplementary video 1). In addition, we repeated the test with different granular stalk lengths. The results show that a longer granular stalk would improve the ultimate adaptation angle and force. For example, for 10mm, the force increased to a range of 0.43N at 80$^\circ$ to 0.69N at 45$^\circ$, while for 5mm, the force was 0.51N-1.31N. During the test, the pressure of the suction cup was measured. As presented in Fig \ref{fig:pressure_vs_force}, the pressure inside the granular stalk is -8KPa which is due to small self jamming between granular particles, and -60KPa after attachment to the surface. For all angles the response time for the suction cup to reach vacuum pressure was around 0.4 second, and 3 seconds for vacuum release during the detachment.

Furthermore, we changed the granular stalk to silicone rubber and repeated the experiment to compare the adaptation of the proposed design with using only material softness. For the Ecoflex 00-10 stalk, the suction cup adapted to angles up to 65$^\circ$ with a force range of 0.51N at 15$^\circ$ to 1.15N at 45$^\circ$  (See Supplementary Video 1). For Dragonskin 10, the adaptation force increased significantly to almost 5N at a 45$^\circ$ angle which was also the ultimate adaptation.   

In order to investigate the effect of the suction pad stiffness on adaptation, we changed the suction pad from Dragonskin 10 to Ecoflex 00-10 and attached the pad to 20mm granular stalk. With this change, the measured force at 15$^\circ$ was 0.36N, lower than 20mm granular with Dragonskin pad. However, at 30$^\circ$, the force rises to 0.73N which is higher than granular cases with Dragonskin pad. Moreover, in this case, the ultimate adaptation angle was only 45$^\circ$, much lower than 20mm coffee with stiffer pad.  

\begin{table}
\centering
\caption{Ultimate adaptation angle and the required force for adaptation at these angles for various scenarios}
\renewcommand{\arraystretch}{1.2}
\begin{tabular}{ |c|c|c| } 
\hline
 Adaptation  & Ultimate & Force at  \\
  Scenarios & adaptation angles ($^\circ$) & ultimate angle (N) \\[0.5ex]
 \hline
  20mm Granular & 85 & 0.33 \\ 
  \hline
 10mm Granular & 80 & 0.43 \\
 \hline
 5mm Granular & 55 & 1.17 \\
 \hline
 Ecoflex 00-10 & 70 & 1.09 \\
 \hline
 Dragonskin 10 & 45 & 4.96 \\
 \hline
 Ecoflex 00-10 suction pad & 45 & 0.69 \\
 \hline
\end{tabular}
\renewcommand{\arraystretch}{1}
\label{table}
\end{table}

\subsection{Bending stiffness test}
\begin{figure}
\centerline
{\includegraphics[width=1\textwidth]{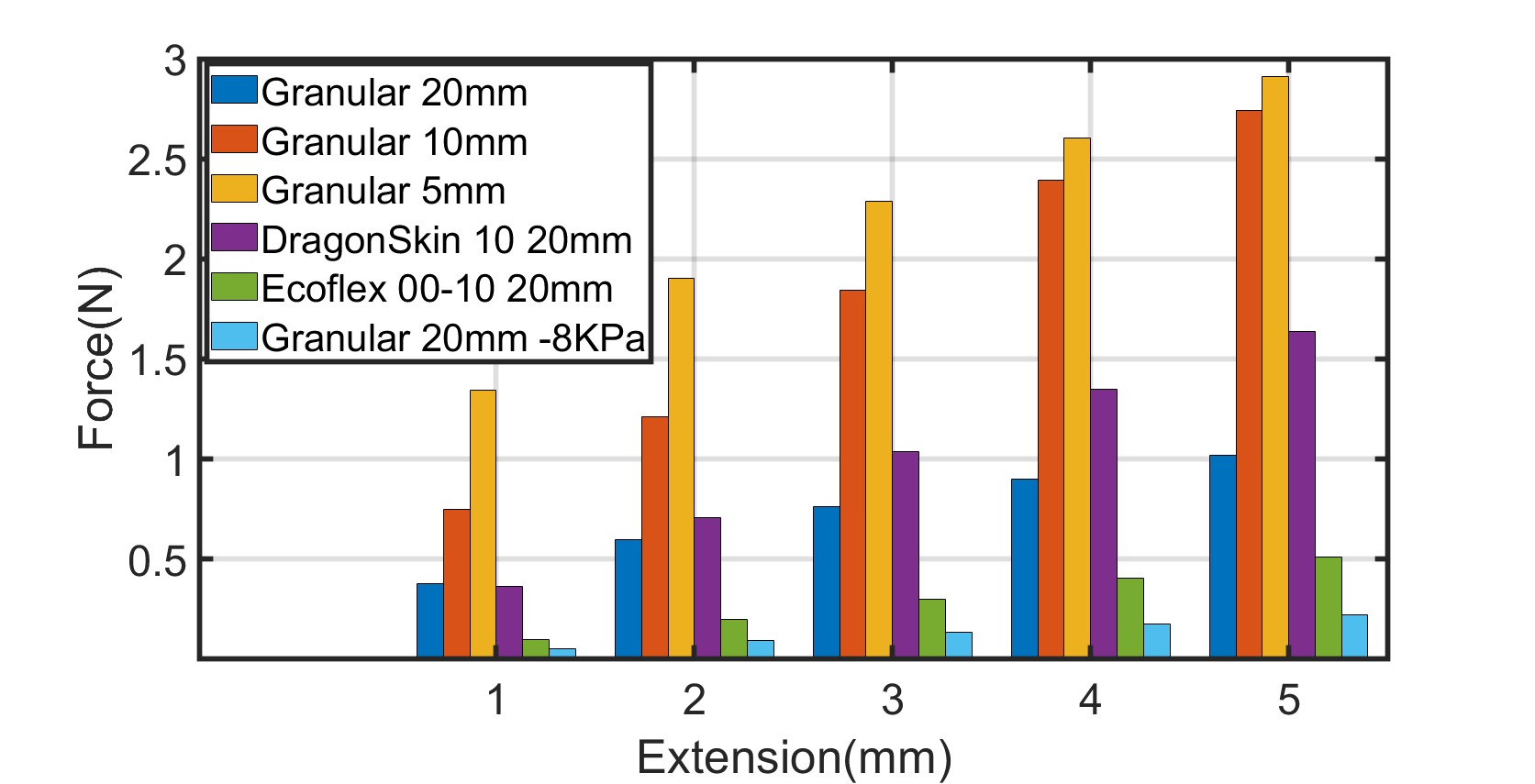}}
\caption{The required force to deflect the stalk's end in various scenarios and the deflection values. The pressure for all scenarios except the last one is -60KPa. In the last scenario the pressure was set to -8KPa to measure the bending stiffness of the stalk due to granular particles self jamming before surface attachment.}
\label{fig:bending}
\end{figure}

As mentioned in \ref{experimental_setup}, we conducted the bending stiffness test to investigate the granular jamming impact on the stalk's bending stiffness and compare it to silicone rubber stalks. Here, the force was applied to the stalk's end and measured until the stalk end deflected 5mm. Looking at Fig \ref{fig:bending}, it is apparent that shorter granular joints improve the stalk's shear force strength. For instance, 1.02N deflected 20mm granular stalk 5mm, while the required for such deflection for 10mm and 5mm were 2.74N and 2.91N, respectively.  

Moreover, the obtained results show that using granular jamming instead of uniformly soft stalk provides higher bending stiffness. For example, the measured force at 5mm deflection for Ecoflex 00-10 stalk was 0.51N, lower than all the granular scenarios. For Dragonskin 10 also, the maximum recorded shear force was 1.64N, which means 10mm and 5mm granular stalks can resist higher shear force than Dragonskin10.

\section{Discussion}

\begin{figure}
    \centering
    \includegraphics[width=0.6\textwidth]{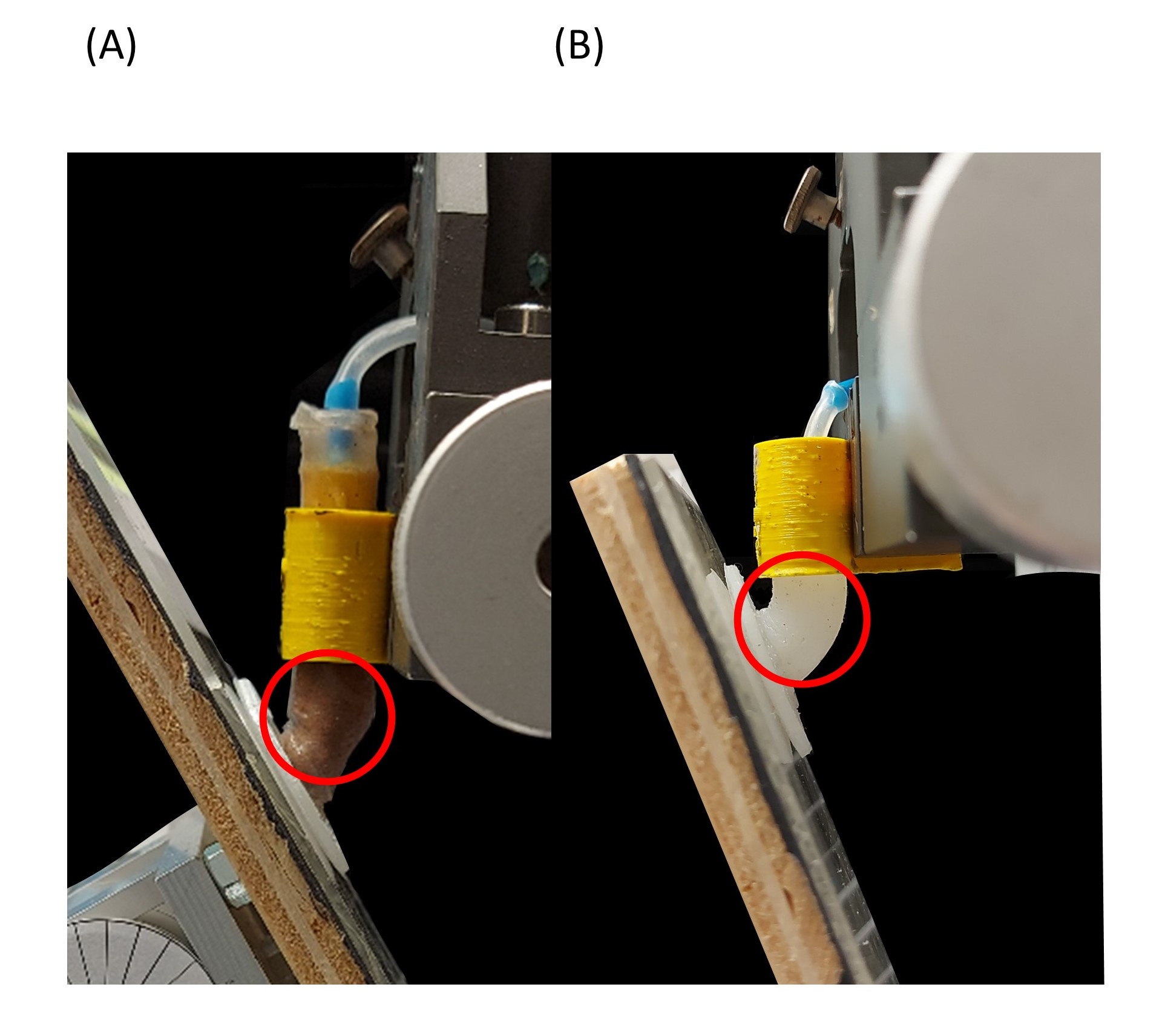}
    \caption{The suction cup shape while attached to a steep angle. (A) the wrinkle and hinge-like behavior of the granular stalk (B) the uniform bending curvature in silicone rubber stalk.}
    \label{fig:shape}
\end{figure}

In this letter, we describe a suction cup design capable of adapting angular surfaces and assess how stiffness impacts the design in such adaptation. The results from the conducted experiments indicate that granular stalks adapting to angular surfaces required less force (max 0.48N) than silicone rubber stalks with the same length (max 1.15N for Ecoflex and 4.96N for Dragonskin) due to the lower stiffness of granular particles without jamming as. In addition, the Ecoflex stalk adapted easier to higher angles than the Dragonskin stalk. Therefore, these results agree with Eq \ref{eq:2nd}, in which with having same $\alpha$, more compliant stalk due to lower Young's modulus requires less adaptation force. Besides low stiffness, another reason for the significant difference in the adaptation of the granular stalk is that the granular particles reshape during adaptation. As shown in Fig \ref{fig:shape}, such reshaping endows the granular stalk with a hinge-like behavior to adapt easier to steep angles, unlike silicone rubber stalk, which requires uniform bending to achieve adaptation. These results suggest that using granular stalks perform better adaptation than silicone rubber stalks. 

Furthermore, the adaptation force and the ultimate angle were compared for various granular stalk lengths. The results reveal that a longer stalk requires less force to adapt to angular surfaces and adapts to higher angles because of higher compliance. In addition to compliance, another possible explanation for this result is the suction cup geometry during adaptation to steep angles. For instance, with a 20mm diameter suction pad, the 5mm granular stalk was too short of providing enough bending motion required for adaptation to angles higher than 60$^\circ$, and the force increased remarkably at 45$^\circ$.

Another point in the adaptation test is the trend that emerged in which for granular stalks and Ecoflex stalk extremum occurs in the recorded forces. Although from Eq\ref{eq:2nd}, we expect higher forces at higher angles, but during the adaptation, the stalks would wrinkle and behave similar to a buckled beam which result in lower forces at higher angles. As expected, the extremum angle is higher for the stiffer stalk, since the stiffer stalk requires more force to be wrinkled and behave as a buckled beam. For instance, the maximum for 20mm was recorded at 30$^\circ$, while for 10mm and 5mm, it was at 45$^\circ$.

By combining the adaptation and stiffness experiment results, the 20mm granular suction cup not only adapts with less than 0.5N force to angles up to 85$^\circ$ but also had twice the bending stiffness of Ecoflex 00-10 stalk, which required twice the force at some angles. In addition, 10mm and 5mm granular stalks had higher bending stiffness than Dragonskin 10 stalk, which required nearly seven times more force adapting to 45$^\circ$ surface than 10mm granular. Hence, these findings suggest that granular stalks endow the suction cup with more effortless passive adaptation to angular surfaces and provide higher bending stiffness than passive adaptation with just material flexibility. 

\begin{figure}
    \centering
    \includegraphics[width=0.8\textwidth]{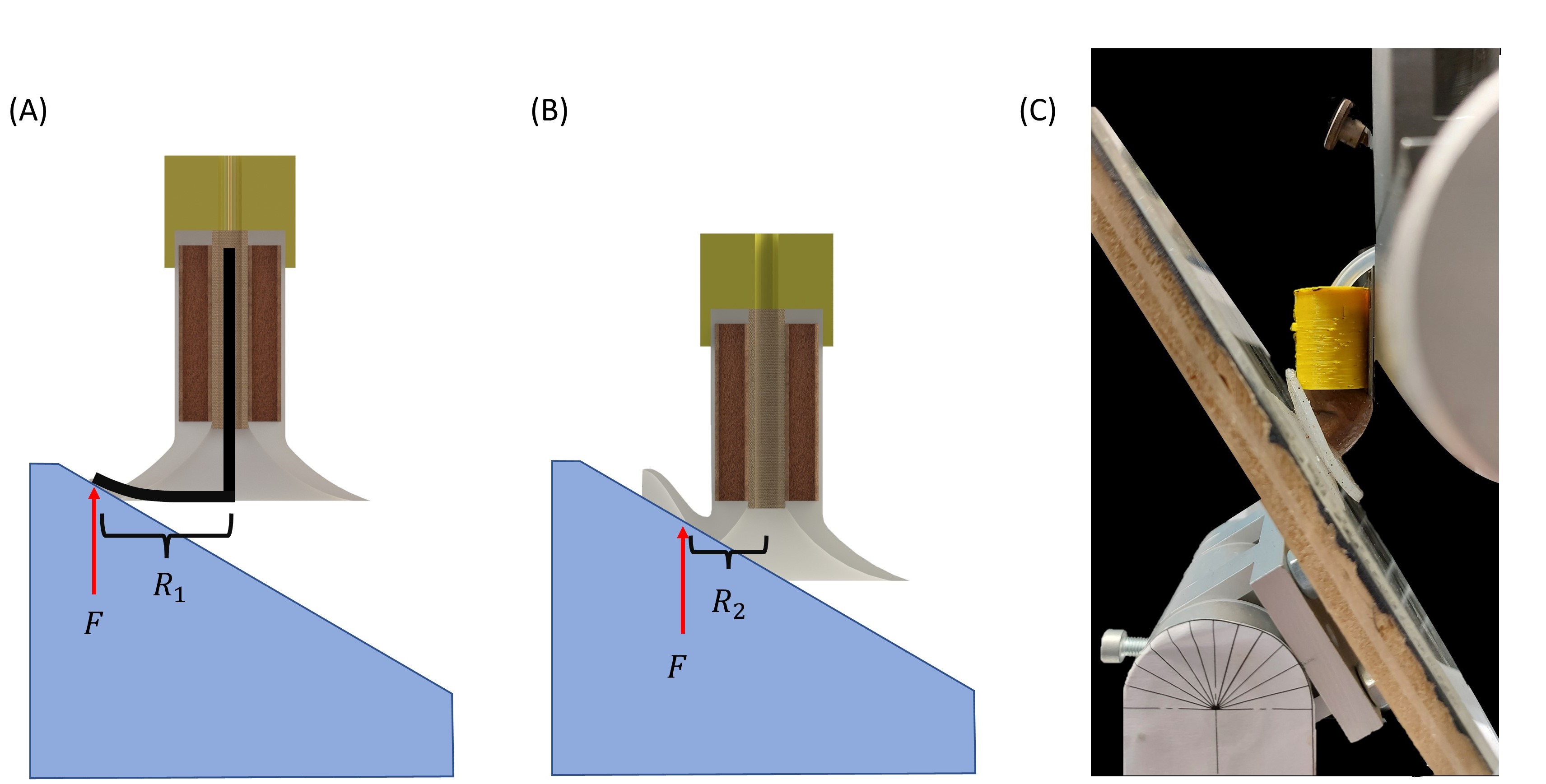}
    \caption{ (A) Deformation of suction cup with Dragonskin pad which keep the moment arm ($R_{1}$) enough for bending the stalk. (B) Deformation of the suction cup with Ecoflex 00-10 suction pad, with shorter moment arm ($R_{2}$) due to flexibility of the pad. (C) The adaptation of suction cup with Ecoflex suction pad which deformed irregularly compare to Dragonskin pad.}
    \label{fig:ecopad}
\end{figure}

In another experiment, one unanticipated result occurred when we replaced the Dragonskin suction pad with Ecoflex. We expected using more compliant materials for the suction pad decreases the adaptation force, which also can be seen in the result at 15$^\circ$. However, at 30$^\circ$ and 45$^\circ$, the suction cup with Ecoflex pad required more adaptation force than the Dragonskin pad. This unexpected result is because the softer suction pad deforms more and faster than the stiffer pad which leads to moment arm of L-shape beam be smaller Fig\ref{fig:ecopad}. As a result, the moment on the stalk will be lower and makes the adaptation harder at higher angles. In addition, shortening the moment arm also leads to significant deformation in the stalk (Fig\ref{fig:ecopad}) which also increase the adaptation force.
Moreover, the ultimate adaptation angle of this suction cup was only 45$^\circ$, much lower than the Dragonskin pad. Therefore, although Ecoflex suction pads can attach to rougher surfaces, using such pad impacts the adaptation drastically. It is important to bear in mind that using a material with more stiffness than Dragonskin does not necessarily improve the adaptation result, since stiffer suction pad may require much higher force for deformation and may not changing the moment arm considerably. Further studies are needed to understand better how the suction pad design itself affects the adaptation.

For final discussion, although in theory, we can calculate the required force for adaptation at any angle, in reality, if the force overcomes the friction between the suction pad and the surface, the suction cup slips on the surface. In such situations, the applied force only moves the suction pad on the surface and can not bend the stalk to achieve adaptation. With this point in mind, it is also possible to calculate the ultimate adaptation angle, which is beyond the scope of this study. Furthermore, despite the theory shows the general trend of the suction cup behaviour, there is an error between the actual measured force and the force obtained from the theory. The main reason behind the difference is that the suction pad with the stalk is modelled as a single L-shape beam while in reality there are deformation at the connection point which is neglected in the model. In addition, we represent the stalk as a beam with uniform young's modulus. However, due to composite structure of the granular stalk, and also particles movement during adaptation, this assumption is not accurate as well.

\section{Conclusion and Future Work}
This study sets out to develop a suction cup design capable of adapting to angular surfaces and having high lateral stiffness after adaptation. The results of this study have identified that using granular jamming techniques offers adaptation and lateral stiffness, which could not be matched with adaptation with only material flexibility. Therefore, the proposed design could be used to enhance the performance in unconstructed areas. For instance, by having a 10mm granular suction cup as the adhesion to surface method, the actuator or the robot can attach to all surfaces with less than 80$^\circ$ slope, with less than 0.7N, which most soft actuators can provide. Furthermore, the suction cup would tolerate 270gr weight with only 5mm deflection. Besides adaptation to high angles and high lateral stiffness, one main strength of this design is that the modulation is sensorless and without an additional actuator, which reduces the complexity in both design and control. Therefore, this design allows soft actuators to adapt and manipulate easily in undefined environments.


\begin{figure}
    \centering
    \includegraphics[width=0.8\textwidth]{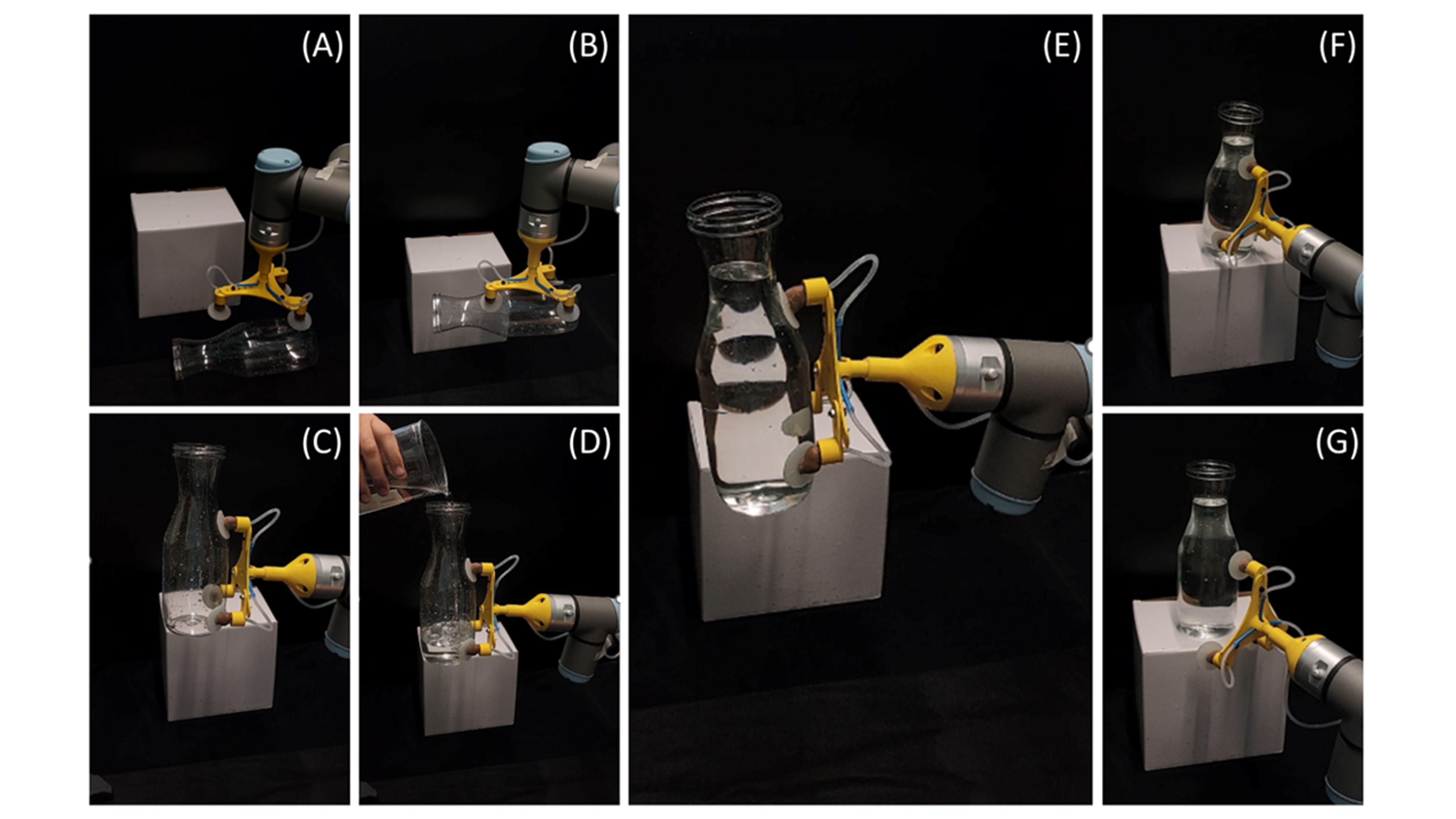}
    \caption{(A)-(C) adaptation to the uneven surfaces of a bottle, (D) adding lateral load by pouring water (E) holding 1028 grams of lateral load (F)-(G) manipulation and releasing the object.}
    \label{fig:application}
\end{figure}

For future work, a more detailed and thorough model would help to optimize the design further. As mentioned in the discussion, the results show that the ratio between the suction pad stiffness and the stalk stiffness considerably influences the suction cup performance. Therefore, the best material stiffness for the suction pad can be selected with a better model. In addition, depending on the actuators and the application, the length of the stalk can accordingly be modified for specific cases. Another interesting future work can be potential applications of this suction cup by adding it to spider robots or manipulating irregular object such as Fig\ref{fig:application} and supplementary video 2.

\bibliographystyle{reference}
\bibliography{reference}

\end{document}